\documentclass[runningheads]{llncs}
\usepackage{amsmath} 
\usepackage{booktabs} 
\usepackage{xcolor}
\usepackage{authblk} 
\usepackage{marvosym}
\usepackage{graphicx}
\usepackage{amssymb}
\usepackage{colortbl}
\usepackage[T1]{fontenc}
\usepackage{multirow}
\usepackage{graphicx}
\usepackage{adjustbox}
\usepackage{graphicx,verbatim}

\usepackage[normalem]{ulem}

\begin{document}

\title{Knowing or Guessing? Robust Medical Visual Question Answering via Joint Consistency and Contrastive Learning}
%

\author{Songtao Jiang\inst{1, 5}\and
Yuxi Chen\inst{1}\and
Sibo Song\inst{2}\and
Yan Zhang\inst{1}\and
Yeying Jin\inst{3}\and
Yang Feng\inst{4}\and
Jian Wu\inst{1,6}\and
Zuozhu Liu\inst{1,6}\textsuperscript{(\Letter)} 
}
\authorrunning{Songtao Jiang et al.}
\institute{Zhejiang University, Zhejiang, China \and
Alibaba Group, Zhejiang, China\and National University of Sinapore, Sinapore
\and
Angelalign Technology Inc., Shanghai, China\and
ChohoTech Inc., Hangzhou, China\and
Zhejiang Key Laboratory of Medical Imaging Artificial Intelligence, Zhejiang, China
\\
}

\maketitle              
\begin{abstract}

In high-stakes medical applications, consistent answering across diverse question phrasings is essential for reliable diagnosis. However, we reveal that current Medical Vision-Language Models (Med-VLMs) exhibit concerning fragility in Medical Visual Question Answering, as their answers fluctuate significantly when faced with semantically equivalent rephrasings of medical questions. We attribute this to two limitations: (1) insufficient alignment of medical concepts, leading to divergent reasoning patterns, and (2) hidden biases in training data that prioritize syntactic shortcuts over semantic understanding.
To address these challenges, we construct RoMed, a dataset built upon original VQA datasets containing 144k questions with variations spanning word-level, sentence-level, and semantic-level perturbations. When evaluating state-of-the-art (SOTA) models like LLaVA-Med on RoMed, we observe alarming performance drops (e.g., a 40\% decline in Recall) compared to original VQA benchmarks, exposing critical robustness gaps.
To bridge this gap, we propose Consistency and Contrastive Learning (CCL), which integrates two key components: (1) knowledge-anchored consistency learning, aligning Med-VLMs with medical knowledge rather than shallow feature patterns, and (2) bias-aware contrastive learning, mitigating data-specific priors through discriminative representation refinement. CCL achieves SOTA performance on three popular VQA benchmarks and notably improves answer consistency by 50\% on the challenging RoMed test set, demonstrating significantly enhanced robustness. Code will be released.

\keywords{ Medical visual question answering \and Medical vision-language models \and Robustness and trustworthiness.}

\end{abstract}

\section{Introduction}
\begin{figure}[t!]
    \centering
    \includegraphics[width=1\linewidth]{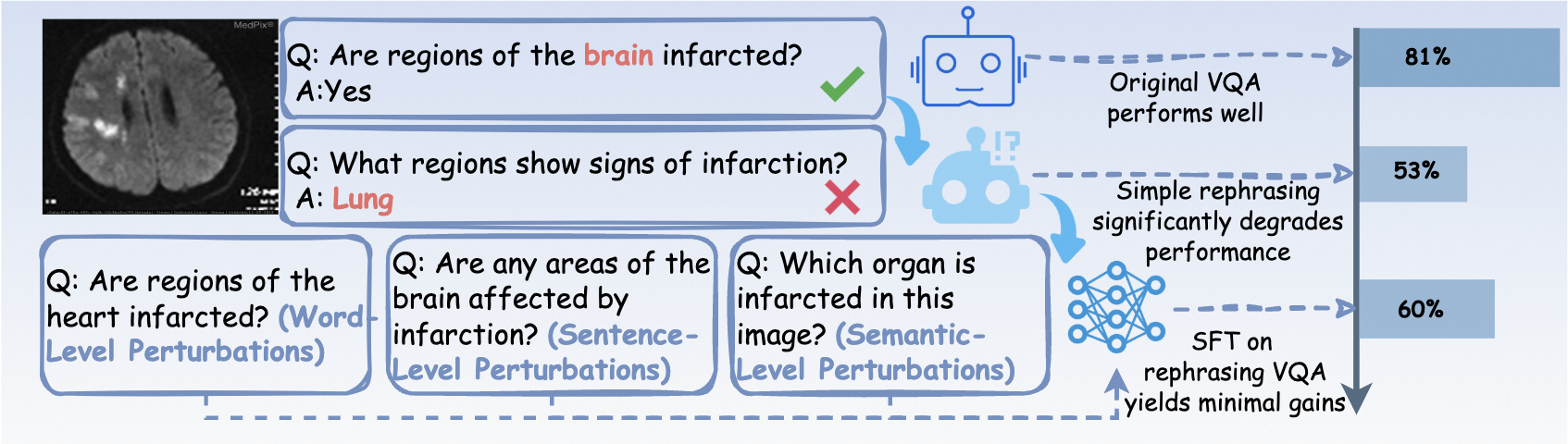}
\caption{A simple perturbation experiment demonstrates that current Med-VLMs exhibit inconsistencies in VQA tasks, raising concerns about the robustness of Med-VQA.}
    \label{fig:intro}
\end{figure}

Recent advancements in Medical Vision-Language Models (Med-VLMs), such as Med-Flamingo~\cite{moor2023med}, Med-PaLM M~\cite{singhal2023towards}, and LLaVA-Med~\cite{li2024llava}, have demonstrated remarkable progress in Medical Visual Question Answering (Med-VQA)~\cite{jiang2024med,jiang2025hscr,jiang2025omniv,jiang2025capo,liu2024medcot}. Through supervised fine-tuning (SFT) on Med-VQA training sets, these models achieve strong performance on downstream tasks. However, as illustrated in Fig.~\ref{fig:intro}, our preliminary tests reveal a critical limitation: When questions are perturbed with varying levels of modifications while preserving semantic equivalence, models often produce inconsistent answers. This inconsistency severely restricts their applicability in real-world clinical settings, where diverse and interactive query formulations are common. Moreover, it raises fundamental concerns about current Med-VQA evaluation framework: \textit{Is the model truly knowing the answers, or is it merely memorizing response patterns and guessing correctly by chance?}

To investigate further, we augmented the diversity of the original training data by introducing word-level perturbations. As shown in Fig.~\ref{fig:intro}, vanilla SFT with more varied training data provides only marginal improvements in robustness against perturbations, with performance still significantly deviating from the original evaluation results. This yields two key insights: (1) the lack of diversity in training data contributes to the inconsistency issue~\cite{shah2019cycle，jiang2024joint,jiang2024modality}, and increasing diversity can mildly mitigate it~\cite{ray2019sunny}; and (2) the current SFT paradigm, with its single autoregressive objective, has inherent limitations, as increasing data diversity alone provides minimal robustness gains. These findings highlight the need for a more robust Med-VQA evaluation framework and training methodology.

\begin{figure}[t!]
    \centering
    \renewcommand{\arraystretch}{0.1}
    \includegraphics[width=0.9\linewidth]{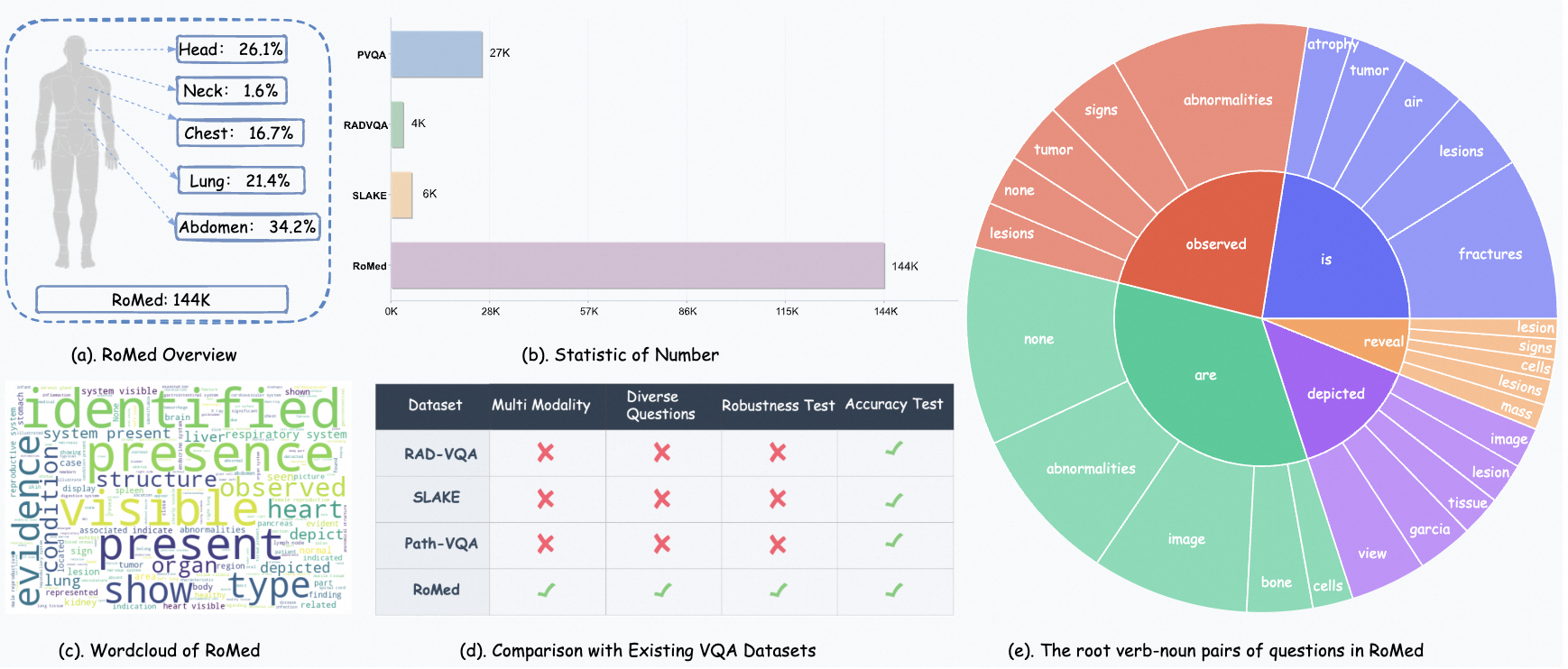}
\caption{Overview of RoMed. RoMed is a comprehensive VQA dataset spanning diverse organs and modalities (CT, MRI, X-Ray, Pathology), with dual evaluations for accuracy and robustness ensuring a holistic assessment.}
    \label{fig:data_overiew}
\end{figure}

To address these challenges, we first construct the RoMed dataset as shown in Fig.~\ref{fig:data_overiew}, a new Med-VQA dataset encompassing training and testing sets across four major medical modalities: CT, MRI, X-Ray, and Pathology. For the training set, we enhance diversity by introducing multi-level perturbations at the word, sentence, and semantic levels, enriching the original Med-VQA training data. For the test set, we reconstruct a more comprehensive VQA benchmark based on mainstream Med-VQA datasets. Unlike traditional datasets~\cite{zhang2023pmc,zhang2023biomedgpt} that focus solely on accuracy, we incorporate evaluation metrics such as the Coefficient of Variation (CV) and Mean Absolute Deviation (MAD) to assess answer consistency, providing a more robust evaluation framework. 

Furthermore, we propose Joint Consistency and Contrastive Learning (CCL) to address the limitations of the current SFT paradigm. Through consistency learning~\cite{Dwibedi_2019_CVPR}, CCL provides explicit supervision signals to ensure the model delivers correct answers across various perturbations, fostering better alignment with medical knowledge rather than shallow, overfitting features. Additionally, by treating perturbed questions as positive samples and using other questions in the same batch as negative samples, CCL guides the model to perform comparative understanding by leveraging contrastive learning objective~\cite{khosla2020supervised}. This dual-objective approach mitigates potential overfitting in the model’s representations and significantly enhances its generalization capabilities, making it more robust and reliable for real-world clinical applications. Extensive experiments and analyses demonstrate that CCL not only significantly enhances Med-VQA performance but also markedly reduces MAD and CV, thereby improving model robustness. CCL achieves state-of-the-art (SoTA) accuracy and robustness on widely-used benchmarks, including Rad-VQA~\cite{lau2018dataset}, SLAKE~\cite{liu2021slake}, and PathVQA~\cite{he2020pathvqa}.

\begin{figure}[t!]
    \centering
    \includegraphics[width=1\linewidth]{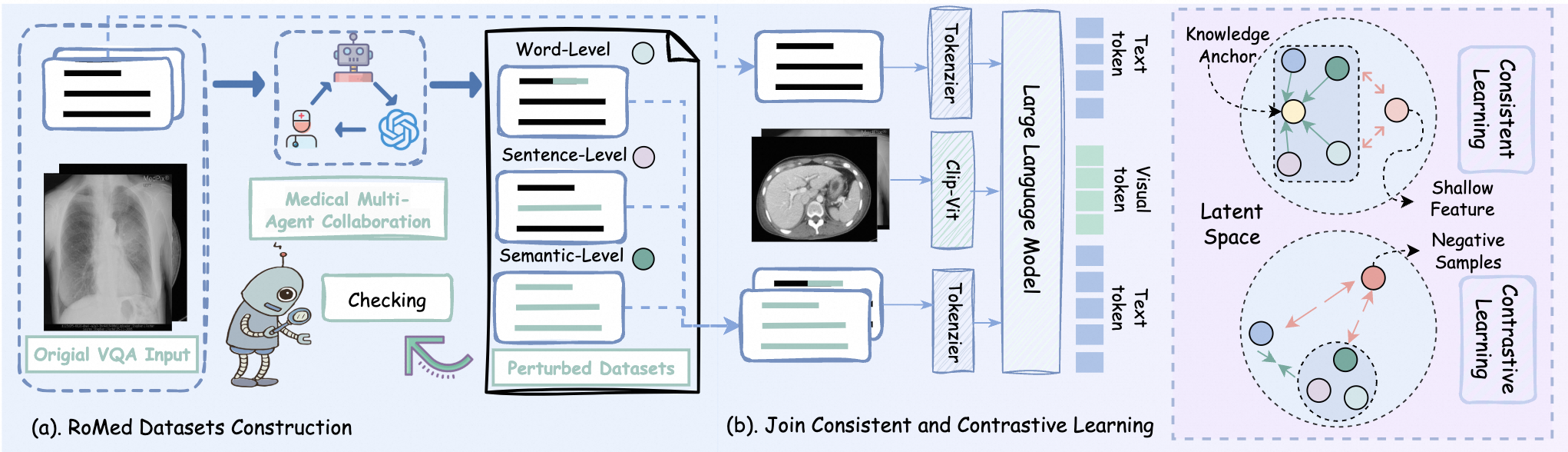}
\caption{Overview of our CCL pipeline. Our framework consists of two key components: (a) constructing the RoMed dataset through medical multi-agent collaboration; (b) joint knowledge-anchored consistency learning for medical expertise alignment and bias-aware contrastive learning to reduce inherent representation biases.}
    \label{fig:pipline}
\end{figure}

\section{Methods}
In this section, we first describe the construction of the RoMed dataset, addressing the lack of robustness evaluation in current Med-VQA systems. To tackle the limited generalization of vanilla SFT, we introduce our Joint Consistency and Contrastive Learning (CCL) framework, which optimizes Med-VLM representation learning by integrating consistency and contrastive objectives.

\subsection{RoMed Datasets Construction}
Our study reveals that current Med-VQA systems often fail to answer semantically equivalent perturbed questions correctly {(see Fig.~\ref{fig:intro})}, despite accurately answering the original questions. This suggests that the reported accuracy on existing Med-VQA benchmarks may not reliably reflect the true knowledge level of Med-VLMs. To address this limitation, we construct a more diverse and robust evaluation dataset for Med-VQA (see Fig.~\ref{fig:pipline} a).
First, we integrate widely used Med-VQA datasets, including Rad-VQA~\cite{lau2018dataset}, SLAKE~\cite{liu2021slake}, and PathVQA~\cite{he2020pathvqa}, which cover various organs and modalities. Building on these datasets, we introduce perturbations at three levels: word-level, sentence-level, and semantic-level, using a medical multi-agent collaboration system. Specifically, we leverage three models to generate high-quality perturbations, combining both general-purpose and domain-specific VLMs. For the medical multimodal agent, we employ HuatuoGPT-Vision-34B~\cite{chen2024huatuogptvisioninjectingmedicalvisual}, a leading medical VLM, which provides domain-specific medical knowledge by generating captions for the given medical images. For the medical reasoning agent, we use HuatuoGPT-o1~\cite{chen2024huatuogpto1medicalcomplexreasoning}, a single-modal LLM with advanced reasoning capabilities. This agent takes the captions and question-answer pairs as input to produce intermediate reasoning steps for sampling correct reasoning processes. Finally, we utilize GPT-4o as the general meta-agent, a state-of-the-art closed-source model, to integrate feedback from both the medical multimodal and reasoning agents, generating three levels of perturbed questions along with their corresponding answers. After this process, we validate the constructed questions by feeding them back to GPT-4o, ensuring they align with the same medical knowledge as the original questions and do not require additional knowledge beyond what is needed to answer the original questions correctly.
Following this validation step, we construct the RoMed dataset, as shown in Fig.~\ref{fig:data_overiew}.  Since perturbed questions are derived from the original ones, an ideal robust VLM should consistently answer all variants correctly within each question cluster. To quantify consistency, we introduce two metrics: Mean Absolute Deviation (MAD), defined as \(\text{MAD} = \frac{1}{N} \sum_{i=1}^{N} |x_i - \mu|\), and Coefficient of Variation (CV), defined as \(\text{CV} = \frac{\sigma}{\mu} \times 100\%\), where \(N\) is the number of questions in a cluster, \(x_i\) is the model's answer to the \(i\)-th question, \(\mu\) is the mean of the answers, and \(\sigma\) is the standard deviation.

\subsection{Joint Consistency and Contrastive Learning}
\subsubsection{Knowledge-Anchored Consistency Learning}
Let $q$ denote the original question, tokenized into text tokens $\mathcal{T}_q$. The corresponding image $I$ is encoded into visual tokens $\mathcal{V}_I$ using a visual encoder (e.g., CLIP-ViT). The multimodal input $\mathcal{X}$ is formed by concatenating $\mathcal{T}_q$ and $\mathcal{V}_I$, i.e., $\mathcal{X} = [\mathcal{T}_q; \mathcal{V}_I]$. The input $\mathcal{X}$ is fed into the LLM backbone, generating outputs $\mathcal{Y}$. The autoregressive loss $\mathcal{L}_{\text{original}}$ is computed as: \(\mathcal{L}_{\text{original}} = -\sum_{t=1}^{T} \log P(y_t \mid y_{<t}, \mathcal{X}),
\)
where $T$ is the output sequence length and $y_t$ is the token at position $t$.
To enhance the alignment with medical knowledge, we perform consistency learning by introducing perturbations at three levels: word-level, sentence-level, and semantic-level. These perturbations are constructed through multi-agent collaboration based on the original question. For each perturbed question $q_i$ ($i \in \{w, s, \text{sem}\}$), the perturbed input $\mathcal{X}_i = [\mathcal{T}_{q_i}; \mathcal{V}_I]$ is used to compute the total consistency loss:
\[
\mathcal{L}_{\text{consistency}} = \mathcal{L}_{\text{original}} + \sum_{i \in \{w, s, \text{sem}\}} \left( -\sum_{t=1}^{T} \log P(y_t \mid y_{<t}, \mathcal{X}_i) \right).
\]
\subsubsection{Bias-Aware Contrastive Learning}

To eliminate bias in the training data and calibrate the model's representation, we employ contrastive learning as part of the CCL framework. Specifically, the original question $q$ and its perturbed versions at three levels ($q_w$, $q_s$, and $q_{\text{sem}}$) are treated as positive samples, while other questions in the same batch serve as negative samples.
The hidden state embedding $\mathcal{H}$ for the original input is obtained by feeding the multimodal input $\mathcal{X} = [\mathcal{T}_q; \mathcal{V}_I]$ into the LLM backbone and applying mean pooling: $\mathcal{H} = \mathcal{M}(\mathcal{LLM}(\mathcal{X}))$, where $\mathcal{M}(\cdot)$ denotes the mean pooling operation. Similarly, for each perturbed question $q_p$ (with $p \in \{w, s, \text{sem}\}$), the corresponding hidden state embedding is $\mathcal{H}_p^+ = \mathcal{M}(\mathcal{LLM}([\mathcal{T}_{q_p}; \mathcal{V}_I]))$. The embeddings of unrelated questions in the batch are denoted as $\mathcal{H}_j^-$.
The contrastive loss $\mathcal{L}_{\text{ctr}}$ is:
\[
\mathcal{L}_{\text{ctr}} = -\sum_{p \in \{w, s, \text{sem}\}} \log \frac{\exp(\text{sim}(\mathcal{H}, \mathcal{H}_p^+) / \tau)}{\exp(\text{sim}(\mathcal{H}, \mathcal{H}_p^+) / \tau) + \sum_{j=1}^{N} \exp(\text{sim}(\mathcal{H}, \mathcal{H}_j^-) / \tau)},
\]
where $\mathcal{H}_j^-$ represent the embedding of the $j$-th negative sample, $\text{sim}(\cdot, \cdot)$ denote the cosine similarity, $\tau > 0$ be a temperature hyperparameter, and $N$ is the total number of negative samples. The overall loss is defined as $\mathcal{L} = \frac{\mathcal{L}_{\text{ctr}} + \mathcal{L}_{\text{consistency}}}{2}$.

\begin{table}[ht!]
\centering
\renewcommand{\arraystretch}{0.1}
\resizebox{0.8\textwidth}{!}{
\begin{tabular}{llcc|cc|cc}  
\toprule
\textbf{Method} & & \multicolumn{2}{c|}{\cellcolor[HTML]{ffffff}\textbf{RAD-VQA}} & \multicolumn{2}{c|}{\cellcolor[HTML]{ffffff}\textbf{SLAKE}} & \multicolumn{2}{c}{\cellcolor[HTML]{ffffff}\textbf{PathVQA}} \\
 & & \cellcolor[HTML]{ffffff}\textbf{Open} & \cellcolor[HTML]{ffffff}\textbf{Closed} & \cellcolor[HTML]{ffffff}\textbf{Open} & \cellcolor[HTML]{ffffff}\textbf{Closed} & \cellcolor[HTML]{ffffff}\textbf{Open} & \cellcolor[HTML]{ffffff}\textbf{Closed} \\
\midrule
\rowcolor[HTML]{F3F4F6}
\multicolumn{8}{l}{\textit{Representative \& SoTA methods reported in the literature (Non-VLMs Based Methods)}} \\
\midrule
VL Encoder--Decoder~\cite{bazi2023vision} & & \cellcolor[HTML]{ffffff}- & \cellcolor[HTML]{ffffff}82.5 & \cellcolor[HTML]{ffffff}- & \cellcolor[HTML]{ffffff}- & \cellcolor[HTML]{ffffff}- & \cellcolor[HTML]{ffffff}85.6 \\
Q2ATransformer~\cite{liu2023q2atransformer} & & \cellcolor[HTML]{ffffff}- & \cellcolor[HTML]{ffffff}81.2 & \cellcolor[HTML]{ffffff}- & \cellcolor[HTML]{ffffff}- & \cellcolor[HTML]{ffffff}54.9 & \cellcolor[HTML]{ffffff}88.9 \\
Prefix T. Medical LM~\cite{van2023open} & & \cellcolor[HTML]{ffffff}- & \cellcolor[HTML]{ffffff}- & \cellcolor[HTML]{ffffff}- & \cellcolor[HTML]{ffffff}82.0 & \cellcolor[HTML]{ffffff}- & \cellcolor[HTML]{ffffff}87.0 \\
PubMedCLIP~\cite{eslami2023pubmedclip} & & \cellcolor[HTML]{ffffff}- & \cellcolor[HTML]{ffffff}80.0 & \cellcolor[HTML]{ffffff}- & \cellcolor[HTML]{ffffff}82.5 & \cellcolor[HTML]{ffffff}- & \cellcolor[HTML]{ffffff}- \\
BiomedCLIP~\cite{zhang2023large} & & \cellcolor[HTML]{ffffff}- & \cellcolor[HTML]{ffffff}79.8 & \cellcolor[HTML]{ffffff}- & \cellcolor[HTML]{ffffff}89.7 & \cellcolor[HTML]{ffffff}- & \cellcolor[HTML]{ffffff}- \\
M2I2~\cite{li2022self} & & \cellcolor[HTML]{ffffff}- & \cellcolor[HTML]{ffffff}83.5 & \cellcolor[HTML]{ffffff}- & \cellcolor[HTML]{ffffff}91.10 & \cellcolor[HTML]{ffffff}- & \cellcolor[HTML]{ffffff}88.0 \\
BiomedGPT-S~\cite{zhang2023biomedgpt} & & \cellcolor[HTML]{ffffff}13.4 & \cellcolor[HTML]{ffffff}57.8  & \cellcolor[HTML]{ffffff}66.5 & \cellcolor[HTML]{ffffff}73.3 & \cellcolor[HTML]{ffffff}10.7 & \cellcolor[HTML]{ffffff}84.2 \\
BiomedGPT-M~\cite{zhang2023biomedgpt} & & \cellcolor[HTML]{ffffff}53.6 & \cellcolor[HTML]{ffffff}65.0 & \cellcolor[HTML]{ffffff}78.3 & \cellcolor[HTML]{ffffff}86.8 & \cellcolor[HTML]{ffffff}12.5 & \cellcolor[HTML]{ffffff}85.7 \\
CLIP-ViT w/ GPT2-XL & & \cellcolor[HTML]{ffffff}- & \cellcolor[HTML]{ffffff}- & \cellcolor[HTML]{ffffff}84.3 & \cellcolor[HTML]{ffffff}82.1 & \cellcolor[HTML]{ffffff}40.0 & \cellcolor[HTML]{ffffff}87.0 \\
\midrule
\rowcolor[HTML]{F3F4F6}
\multicolumn{8}{l}{\textit{VLMs-based results}} \\
\midrule
GPT-4o~\cite{hurst2024gpt} & & \cellcolor[HTML]{ffffff}{51.6} & \cellcolor[HTML]{ffffff}{64.0} & \cellcolor[HTML]{ffffff}59.1 & \cellcolor[HTML]{ffffff}71.6 & \cellcolor[HTML]{ffffff}24.1 & \cellcolor[HTML]{ffffff}76.0 \\
LLaVA-v1.5~\cite{Liu_2024_CVPR} & & \cellcolor[HTML]{ffffff}{23.6} & \cellcolor[HTML]{ffffff}{50.7} & \cellcolor[HTML]{ffffff}35.2 & \cellcolor[HTML]{ffffff}52.2 & \cellcolor[HTML]{ffffff}11.9 & \cellcolor[HTML]{ffffff}52.8 \\
Med-Flamingo~\cite{moor2023med} & & \cellcolor[HTML]{ffffff}10.3 & \cellcolor[HTML]{ffffff}52.2 & \cellcolor[HTML]{ffffff}{8.5} & \cellcolor[HTML]{ffffff}37.0 & \cellcolor[HTML]{ffffff}1.2 & \cellcolor[HTML]{ffffff}45.6 \\
PMC-VQA~\cite{zhang2023pmc} & & \cellcolor[HTML]{ffffff}6.3 & \cellcolor[HTML]{ffffff}41.5 & \cellcolor[HTML]{ffffff}7.3 & \cellcolor[HTML]{ffffff}33.9 & \cellcolor[HTML]{ffffff}{1.0} & \cellcolor[HTML]{ffffff}40.1 \\
SQ-LLaVA~\cite{sun2025sq} & & \cellcolor[HTML]{ffffff}23.9 & \cellcolor[HTML]{ffffff}52.6 & \cellcolor[HTML]{ffffff}{40.0} & \cellcolor[HTML]{ffffff}{57.5} & \cellcolor[HTML]{ffffff}{12.2} & \cellcolor[HTML]{ffffff}53.8 \\
ST-LLaVA~\cite{sun2024stllava} & & \cellcolor[HTML]{ffffff}{33.8} & \cellcolor[HTML]{ffffff}{59.2} & \cellcolor[HTML]{ffffff}{40.1} & \cellcolor[HTML]{ffffff}{55.5} & \cellcolor[HTML]{ffffff}10.4 & \cellcolor[HTML]{ffffff}52.1 \\
LLaVA-Med (StableLM) & & \cellcolor[HTML]{ffffff}51.6 & \cellcolor[HTML]{ffffff}75.4 & \cellcolor[HTML]{ffffff}82.2 & \cellcolor[HTML]{ffffff}82.7 & \cellcolor[HTML]{ffffff}33.2 & \cellcolor[HTML]{ffffff}89.5 \\
LLaVA-Med (StableLM) + CCL & & \cellcolor[HTML]{ffffff}\underline{62.7} & \cellcolor[HTML]{ffffff}\underline{84.9} & \cellcolor[HTML]{ffffff}\underline{83.6} & \cellcolor[HTML]{ffffff}{85.1} & \cellcolor[HTML]{ffffff}\underline{36.3} & \cellcolor[HTML]{ffffff}{90.1} \\
LLaVA-Med (Phi2) & & \cellcolor[HTML]{ffffff}54.5 & \cellcolor[HTML]{ffffff}79.8 & \cellcolor[HTML]{ffffff}82.1 & \cellcolor[HTML]{ffffff}\underline{86.5} & \cellcolor[HTML]{ffffff}34.0 & \cellcolor[HTML]{ffffff}\underline{90.4} \\
LLaVA-Med (Phi2) + CCL & & \cellcolor[HTML]{ffffff}\textbf{65.0} & \cellcolor[HTML]{ffffff}\textbf{88.2} & \cellcolor[HTML]{ffffff}\textbf{83.8} & \cellcolor[HTML]{ffffff}\textbf{88.5} & \cellcolor[HTML]{ffffff}\textbf{37.5} & \cellcolor[HTML]{ffffff}\textbf{90.7} \\
\bottomrule
\end{tabular}
}
\caption{Performance on traditional Med-VQA tasks. \textbf{Bold} denotes the best performance,\underline{underlined} denotes the second-best.}
\label{tab:main_table}
\end{table}

\begin{table}[ht!]
\centering
\renewcommand{\arraystretch}{0.1}
\resizebox{1\textwidth}{!}{
\begin{tabular}{llcccc|cccc|cccc}  
\toprule
\textbf{Method} & & \multicolumn{4}{c|}{\cellcolor[HTML]{ffffff}\textbf{RoMed(RAD-VQA)}} & \multicolumn{4}{c|}{\cellcolor[HTML]{ffffff}\textbf{RoMed(SLAKE)}} & \multicolumn{4}{c}{\cellcolor[HTML]{ffffff}\textbf{RoMed(PathVQA)}} \\
 & & \cellcolor[HTML]{ffffff}\textbf{Recall} & \cellcolor[HTML]{ffffff}\textbf{Acc} & \cellcolor[HTML]{ffffff}\textbf{CV(↓)} & \cellcolor[HTML]{ffffff}\textbf{MAD(↓)} & \cellcolor[HTML]{ffffff}\textbf{Recall} & \cellcolor[HTML]{ffffff}\textbf{Acc} & \cellcolor[HTML]{ffffff}\textbf{CV(↓)} & \cellcolor[HTML]{ffffff}\textbf{MAD(↓)} & \cellcolor[HTML]{ffffff}\textbf{Recall} & \cellcolor[HTML]{ffffff}\textbf{Acc} & \cellcolor[HTML]{ffffff}\textbf{CV(↓)} & \cellcolor[HTML]{ffffff}\textbf{MAD(↓)} \\
\midrule
LLaVA-Med (StableLM) & & \cellcolor[HTML]{ffffff}26.5 & \cellcolor[HTML]{ffffff}61.9 & \cellcolor[HTML]{ffffff}83.9 & \cellcolor[HTML]{ffffff}52.1 & \cellcolor[HTML]{ffffff}52.1 & \cellcolor[HTML]{ffffff}72.1 & \cellcolor[HTML]{ffffff}65.3 & \cellcolor[HTML]{ffffff}51.5 & \cellcolor[HTML]{ffffff}22.3 & \cellcolor[HTML]{ffffff}68.4 & \cellcolor[HTML]{ffffff}96.0 & \cellcolor[HTML]{ffffff}58.6 \\
LLaVA-Med (StableLM) + CCL & & \cellcolor[HTML]{ffffff}\underline{48.1} & \cellcolor[HTML]{ffffff}\underline{79.8} & \cellcolor[HTML]{ffffff}\underline{68.3} & \cellcolor[HTML]{ffffff}\underline{42.5} & \cellcolor[HTML]{ffffff}\underline{70.9} & \cellcolor[HTML]{ffffff}\underline{81.3} & \cellcolor[HTML]{ffffff}\underline{57.6} & \cellcolor[HTML]{ffffff}\underline{37.3} & \cellcolor[HTML]{ffffff}\underline{30.8} & \cellcolor[HTML]{ffffff}\underline{81.3} & \cellcolor[HTML]{ffffff}\underline{67.7} & \cellcolor[HTML]{ffffff}\underline{42.4} \\
LLaVA-Med (Phi2) & & \cellcolor[HTML]{ffffff}35.6 & \cellcolor[HTML]{ffffff}63.9 & \cellcolor[HTML]{ffffff}77.8 & \cellcolor[HTML]{ffffff}55.8 & \cellcolor[HTML]{ffffff}60.1 & \cellcolor[HTML]{ffffff}71.9 & \cellcolor[HTML]{ffffff}60.4 & \cellcolor[HTML]{ffffff}49.0 & \cellcolor[HTML]{ffffff}19.2 & \cellcolor[HTML]{ffffff}64.13 & \cellcolor[HTML]{ffffff}93.0 & \cellcolor[HTML]{ffffff}58.4 \\
LLaVA-Med (Phi2) + CCL & & \cellcolor[HTML]{ffffff}\textbf{54.1} & \cellcolor[HTML]{ffffff}\textbf{81.4} & \cellcolor[HTML]{ffffff}\textbf{63.3} & \cellcolor[HTML]{ffffff}\textbf{40.4} & \cellcolor[HTML]{ffffff}\textbf{70.4} & \cellcolor[HTML]{ffffff}\textbf{82.7} & \cellcolor[HTML]{ffffff}\textbf{54.9} & \cellcolor[HTML]{ffffff}\textbf{35.6} & \cellcolor[HTML]{ffffff}\textbf{32.7} & \cellcolor[HTML]{ffffff}\textbf{82.8} & \cellcolor[HTML]{ffffff}\textbf{66.6} & \cellcolor[HTML]{ffffff}\textbf{41.9} \\
\bottomrule
\end{tabular}
}
\caption{Performance on RoMed VQA. \textbf{Bold} denotes the best performance, \underline{underlined} denotes the second-best. Note that lower values are better for CV and MAD.}
\label{tab2}
\end{table}

\section{Experiments}
\noindent \textbf{Evaluation Datasets and Metrics.}
To validate the effectiveness of our proposed CCL in enhancing traditional VQA performance, we conducted experiments on mainstream Med-VQA datasets, including Rad-VQA~\cite{lau2018dataset}, SLAKE~\cite{liu2021slake}, and PathVQA~\cite{he2020pathvqa}. These datasets cover CT, MRI, Chest-Xray, and Pathology modalities, encompassing both open-ended (free-form answers) and closed-ended (yes/no) question settings. For open-ended questions, we used Recall as the evaluation metric, while Accuracy was employed for closed-ended questions, consistent with prior research. Additionally, to assess the robustness of current Med-VLMs, we utilized our constructed RoMed dataset. Beyond Recall and Accuracy, we introduced MAD and CV coefficients to evaluate the consistency of reasoning, reflecting the robustness of Med-VLMs.

\noindent \textbf{Implementation Details.}
For fair comparison, our hyperparameters align with LLaVA-Med~\cite{li2024llava}. We adopt pretrained CLIP-ViT-Large-Patch14 as the vision encoder and StableLM~\cite{bellagente2024stable} and Phi2~\cite{abdin2024phi} as LLM backbones. A 2-layer MLP is used as the projector, with training runs for 9 epochs with a learning rate of 2e-5 without weight decay and a batch size of 2, using 8 × RTX 3090 GPUs.

\noindent \textbf{Baselines.} 
We compare our method with several strong baselines: 
(1) \textit{CLIP-based methods} (e.g., PubMedCLIP~\cite{eslami2023pubmedclip}), which achieve SOTA performance but are limited by their reliance on candidate words for open-ended questions; 
(2) \textit{Medical foundation models} (e.g., BiomedGPT~\cite{zhang2023biomedgpt}), which leverage generative multimodal pretraining but lack multi-turn dialogue capabilities due to their non-LLM architecture; 
(3) \textit{VLM-based models} (e.g., LLaVA-Med, LLaVA-v1.5~\cite{li2024llava,liu2024improved}), which excel in VQA accuracy and interactive dialogue but prioritize precision over robustness. 
In contrast, our CCL method offers a plug-and-play enhancement for medical models, seamlessly integrating with VLM-based approaches to provide multi-turn dialogue support, improved accuracy, and enhanced robustness, making it ideal for real-world clinical applications.

\noindent \textbf{Traditional VQA Performance Comparison.} 
As shown in Tab.~\ref{tab:main_table}, our CCL method, when integrated with the top-performing LLaVA-Med~\cite{li2024llava}, achieves SOTA performance across three benchmarks. Notably, it excels in challenging open-ended questions, demonstrating its effectiveness as a plug-and-play module for robust VQA in clinical settings.

\noindent \textbf{VQA Robustness Performance Comparison.} 
We evaluated our approach on the RoMed VQA benchmark, which introduces variations to assess robustness under diverse clinical queries. As shown in Tab.~\ref{tab2}, LLaVA-Med's accuracy drops significantly (e.g., RAD-VQA~\cite{lau2018dataset} recall decreases by nearly 50\%). In contrast, with CCL, the model maintains high performance, reducing accuracy loss to within 20\% (Fig.~\ref{fig4}). This highlights the limitations of current VQA frameworks and underscores CCL's ability to enhance both performance and robustness for real-world applications.

\begin{table}[t]
\centering
\renewcommand{\arraystretch}{0.1}
\begin{minipage}{0.48\textwidth} 
\centering
\begin{adjustbox}{width=\textwidth} 
\begin{tabular}{cc|cc|cc|cc}
\toprule
\multirow{2}{*}{\cellcolor[HTML]{FFFFFF}A} & \multirow{2}{*}{\cellcolor[HTML]{FFFFFF}B} & \multicolumn{2}{c}{\cellcolor[HTML]{ffffff}RoMed-radvqa} & \multicolumn{2}{c}{\cellcolor[HTML]{ffffff}RoMed-Slake} & \multicolumn{2}{c}{\cellcolor[HTML]{ffffff}RoMed-Pvqa} \\
\cmidrule(lr){3-4} \cmidrule(lr){5-6} \cmidrule(lr){7-8}
 & & \cellcolor[HTML]{ffffff}Recall & \cellcolor[HTML]{ffffff}Acc & \cellcolor[HTML]{ffffff}Recall & \cellcolor[HTML]{ffffff}Acc & \cellcolor[HTML]{ffffff}Recall & \cellcolor[HTML]{ffffff}Acc \\
\midrule
x & x & \cellcolor[HTML]{ffffff}{35.6} & \cellcolor[HTML]{ffffff}{63.9} & \cellcolor[HTML]{ffffff}{60.1} & \cellcolor[HTML]{ffffff}{71.9} & \cellcolor[HTML]{ffffff}{19.2} & \cellcolor[HTML]{ffffff}{64.1} \\
x & $\checkmark$ & \cellcolor[HTML]{ffffff}{\underline{44.5}} & \cellcolor[HTML]{ffffff}{\underline{74.6}} & \cellcolor[HTML]{ffffff}{\underline{65.1}} & \cellcolor[HTML]{ffffff}{\underline{74.6}} & \cellcolor[HTML]{ffffff}{\underline{24.6}} & \cellcolor[HTML]{ffffff}{\underline{70.0}} \\
$\checkmark$ & x & \cellcolor[HTML]{ffffff}{40.7} & \cellcolor[HTML]{ffffff}{65.0} & \cellcolor[HTML]{ffffff}{62.3} & \cellcolor[HTML]{ffffff}{73.5} & \cellcolor[HTML]{ffffff}{20.3} & \cellcolor[HTML]{ffffff}{64.9} \\
$\checkmark$ & $\checkmark$ & \cellcolor[HTML]{ffffff}{\textbf{54.1}} & \cellcolor[HTML]{ffffff}{\textbf{81.4}} & \cellcolor[HTML]{ffffff}{\textbf{70.4}} & \cellcolor[HTML]{ffffff}{\textbf{82.7}} & \cellcolor[HTML]{ffffff}{\textbf{32.7}} & \cellcolor[HTML]{ffffff}{\textbf{82.8}} \\
\bottomrule
\end{tabular}
\end{adjustbox}
\caption{Ablation on joint learning. \textit{A} denotes the consistency learning, and \textit{B} denotes the contrastive learning.}
\label{tab:ablation}
\end{minipage}
\hfill
\begin{minipage}{0.48\textwidth} 
\centering
\renewcommand{\arraystretch}{0.1}
\begin{adjustbox}{width=\textwidth} 
\begin{tabular}{c|cc|cc|cc}
\toprule
\multirow{2}{*}{Model} & \multicolumn{2}{c}{\cellcolor[HTML]{ffffff}RoMed-radvqa} & \multicolumn{2}{c}{\cellcolor[HTML]{ffffff}RoMed-Slake} & \multicolumn{2}{c}{\cellcolor[HTML]{ffffff}RoMed-Pvqa} \\
\cmidrule(lr){2-3} \cmidrule(lr){4-5} \cmidrule(lr){6-7}
 & \cellcolor[HTML]{ffffff}Recall & \cellcolor[HTML]{ffffff}Acc & \cellcolor[HTML]{ffffff}Recall & \cellcolor[HTML]{ffffff}Acc & \cellcolor[HTML]{ffffff}Recall & \cellcolor[HTML]{ffffff}Acc \\
\midrule
Baseline & \cellcolor[HTML]{ffffff}{35.6} & \cellcolor[HTML]{ffffff}{63.9} & \cellcolor[HTML]{ffffff}{60.1} & \cellcolor[HTML]{ffffff}{71.9} & \cellcolor[HTML]{ffffff}{19.2} & \cellcolor[HTML]{ffffff}{64.1} \\
CCL & \cellcolor[HTML]{ffffff}{\underline{54.1}} & \cellcolor[HTML]{ffffff}{\underline{81.4}} & \cellcolor[HTML]{ffffff}{\underline{70.4}} & \cellcolor[HTML]{ffffff}{\underline{82.7}} & \cellcolor[HTML]{ffffff}{\textbf{32.7}} & \cellcolor[HTML]{ffffff}{\underline{82.8}} \\
CCL\textsuperscript{++} & \cellcolor[HTML]{ffffff}{\textbf{55.2}} & \cellcolor[HTML]{ffffff}{\textbf{82.1}} & \cellcolor[HTML]{ffffff}{\textbf{71.6}} & \cellcolor[HTML]{ffffff}{\textbf{83.1}} & \cellcolor[HTML]{ffffff}\underline{32.4} & \cellcolor[HTML]{ffffff}{\textbf{83.3}} \\
\bottomrule
\end{tabular}
\end{adjustbox}
\caption{Model performance comparison under data scaling using LLaVA-Med (Phi2). The variant \textit{CCL++} indicates training with doubled dataset size.}
\label{tab:data_scaling}
\end{minipage}
\end{table}


\subsection{Ablation and Analysis}
\textbf{Ablation of Joint Learning.} We conducted experiments to validate the complementary roles of consistency learning and contrastive learning in our method. As shown in Tab.~\ref{tab:ablation}, the absence of either loss leads to a performance degradation. Contrastive learning plays a critical role in refining robust representations, while consistency learning ensures the model acquires knowledge across varied question formulations and establishes better alignment with medical knowledge. The combination of both components achieves the optimal performance.

\noindent\textbf{Can SFT Improve VQA Robustness?} To verify that our performance improvements are attributable to CCL rather than additional training data, we compared the performance of LLaVA-Med with CCL and vanilla SFT, both trained on the RoMed trainset. As shown in Fig.~\ref{fig5}, vanilla SFT on a larger dataset fails to effectively enhance model robustness. This demonstrates the effectiveness of CCL, which leverages consistency learning to acquire new knowledge while utilizing contrastive learning to refine representations.

\begin{figure}[t!]
    \centering
    \renewcommand{\arraystretch}{0.1}
    \includegraphics[width=0.9\linewidth]{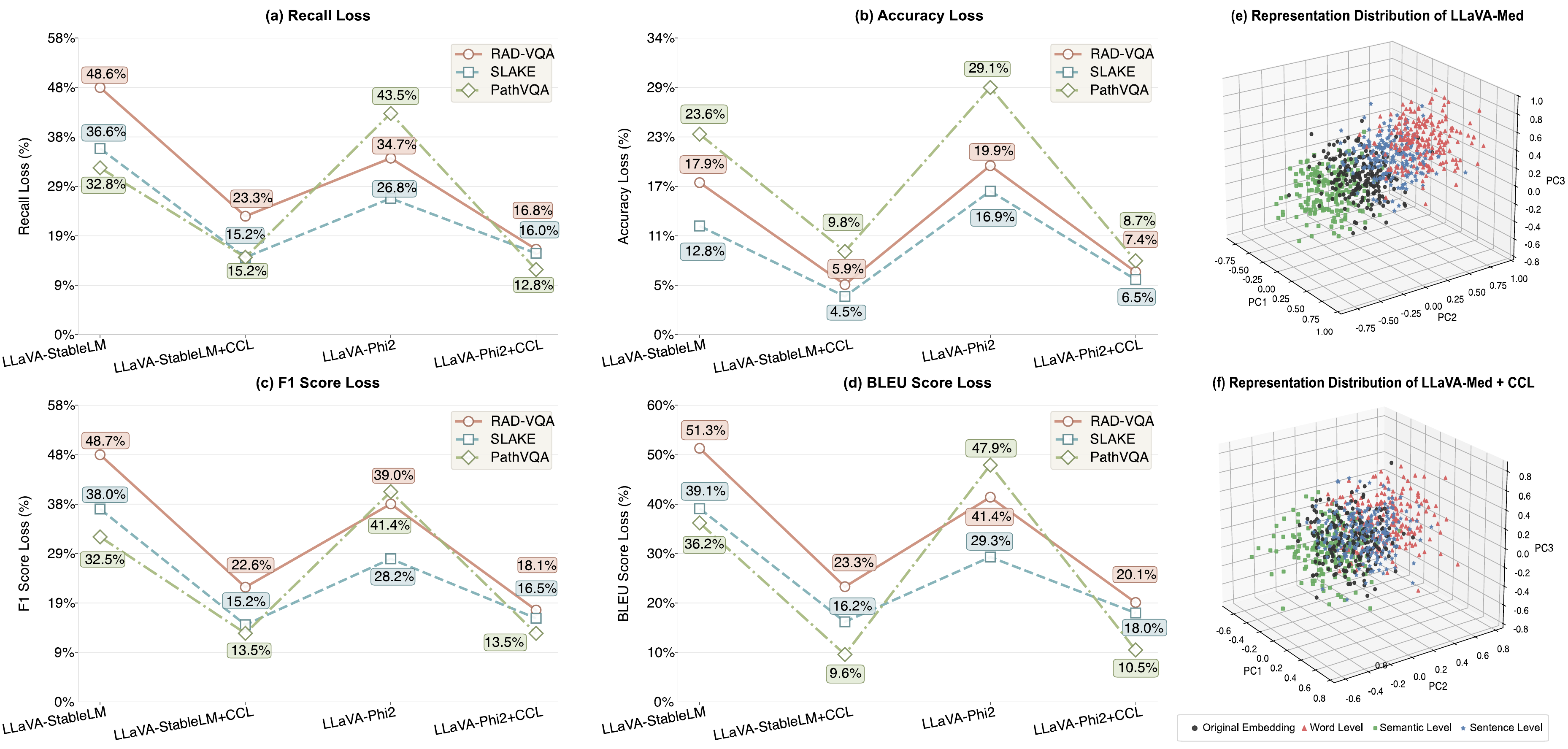}
\caption{(a)--(d) Performance degradation under varied VQA questions, significantly mitigated by CCL; (e)--(f) Representation embeddings of multi-level VQA variations.}
    \label{fig4}
\end{figure}
\begin{figure}[ht!]
    \centering
    \renewcommand{\arraystretch}{0.1}
    \includegraphics[width=0.9\linewidth]{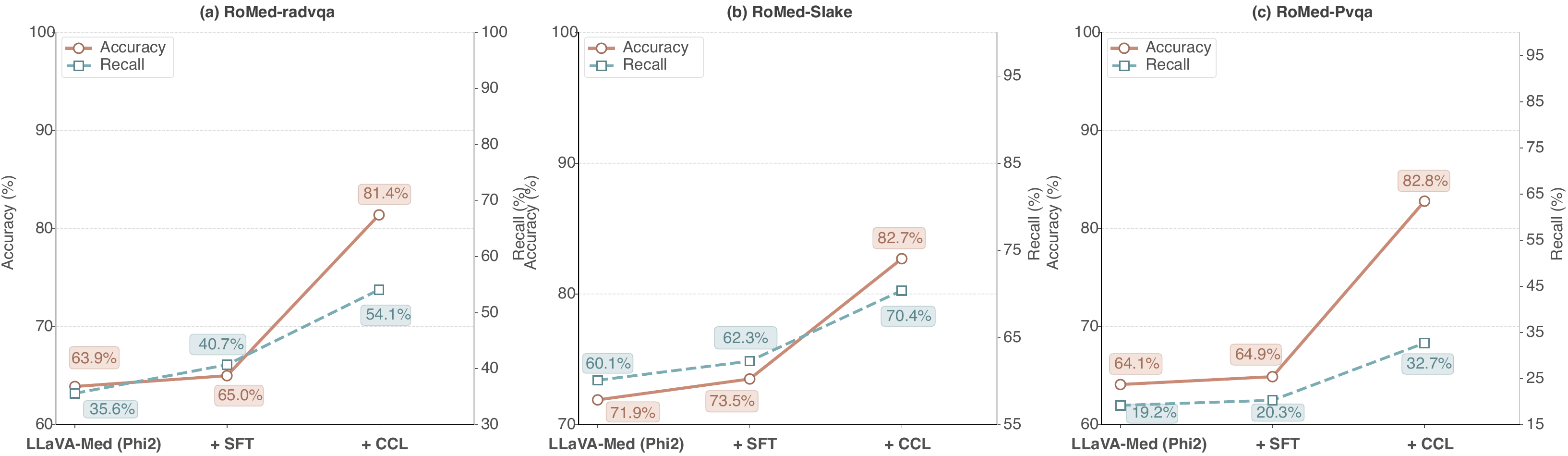}
\caption{Comparison between SFT and CCL. SFT yields minimal performance gains.}
    \label{fig5}
\end{figure}

\noindent\textbf{Representation Visualization Comparison.} As shown in Fig.~\ref{fig4} (e) and (f), we observe that in vanilla LLaVA-Med, the embeddings of the three levels of variations and the original questions are widely separated, indicating that the representations fail to capture the shared features across different formulations. This sensitivity to perturbations could lead to misdiagnoses in real-world clinical applications with diverse query formulations. In contrast, with CCL, the model’s representations under varied perturbations become more robust, suggesting that the model learns more low-level, generalizable features across different levels of perturbations. This makes it better suited for high-stakes clinical scenarios.

\noindent\textbf{Effect of Scaling Data.} To evaluate the effectiveness of our method on larger-scale data, we expanded the original VQA questions by generating two additional variations per level (word-level, sentence-level, and semantic-level), resulting in a dataset twice the size of RoMed training data. This allowed us to explore the trade-off between performance and cost. As shown in Tab.~\ref{tab:data_scaling}, adding one variation per level significantly improves the model's VQA performance and robustness. However, doubling the dataset size yields only marginal gains. Considering the training time overhead, expanding by one variation per level enables the model to achieve strong generalization capabilities through CCL.

\section{Conclusion}
This work reveals the fragility of Med-VLMs in providing consistent answers to semantically equivalent medical questions, attributing it to insufficient concept alignment and training data biases. To address these challenges, we construct RoMed, a dataset with diverse perturbations, and propose Consistency and Contrastive Learning (CCL), which enhances robustness by aligning models with medical knowledge and reducing biases, achieving state-of-the-art performance.
\begin{credits}
\subsubsection{\ackname} This work is supported by the National Natural Science Foundation of China (Grant No. 12326612, 62476241), the Natural Science Foundation of Zhejiang Province, China (Grant No. LZ23F020008), and the Zhejiang University-Angelalign Inc. R\&D Center for Intelligent Healthcare.

\subsubsection{\discintname}
Yang Feng is employed by Angelalign Technology Inc.
\end{credits}

%
%
%
\bibliographystyle{splncs04}
\bibliography{custom}

\end{document}